\documentclass[10pt,journal]{IEEEtran}

\usepackage[T1]{fontenc}
\usepackage[utf8]{inputenc}
\usepackage{microtype}
\usepackage{graphicx}
\usepackage{booktabs}
\usepackage{tabularx}
\usepackage{array}
\usepackage{multirow}
\usepackage{amsmath,amssymb,mathtools,amsthm}
\usepackage{algorithm}
\usepackage{algorithmic}
\usepackage{newfloat}
\usepackage{listings}
\usepackage[table]{xcolor}
\usepackage[caption=false,font=footnotesize]{subfig}
\usepackage[numbers,sort&compress]{natbib}
\usepackage[hyphens]{url}
\usepackage[hidelinks]{hyperref}

\definecolor{lightgray}{gray}{0.75}
\theoremstyle{plain}
\newtheorem{theorem}{Theorem}[section]

\theoremstyle{definition}
\newtheorem{definition}[theorem]{Definition}

\theoremstyle{remark}

\DeclareCaptionStyle{ruled}{labelfont=normalfont,labelsep=colon,strut=off}
\floatstyle{ruled}
\newfloat{listing}{tb}{lst}{}
\floatname{listing}{Listing}

\title{Socialized UAV Cross-Task Learning:\\Towards Cross-Granularity Collaboration through Hierarchical Interaction}

\author{Xinjie Yao\textsuperscript{$\dagger$}, Ruipu Zhao\textsuperscript{$\dagger$}, Yunqi Zhu, Zhihe Fan, Zhoupeng Guo, Weihao Li, Zhen Wang, Qilong Wang, and Pengfei Zhu\textsuperscript{*}
\thanks{
Xinjie Yao is with the Faculty of Information Engineering and Automation, Kunming University of Science and Technology, Kunming 650500, China.
}
\thanks{
Ruipu Zhao, Qilong Wang and Pengfei Zhu are with the School of Artificial Intelligence,
Tianjin University, Tianjin 300350, China.
}
\thanks{
Yunqi Zhu is with the School of Computer Science and Engineering, University of New South Wales, NSW 2052, Australia.
}
\thanks{
Zhihe Fan is with the School of Sports Training, Tianjin University of Sport, Tianjin 300381, China.
}
\thanks{
Zhoupeng Guo is with the School of Automation, Southeast University, Nanjing 210096, China.
}
\thanks{
Weihao Li is with the School of New Media and Communication, Tianjin University, Tianjin 300072, China.
}
\thanks{
Zhen Wang is with the School of Artificial Intelligence, Hebei University of Technology, Tianjin 300401, China.
}
\thanks{
\textsuperscript{$\dagger$} Xinjie Yao and Ruipu Zhao contributed equally to this work.
}
\thanks{
\textsuperscript{*} Corresponding author: Pengfei Zhu.
}
}

\begin{document}

\maketitle

\begin{abstract}
    Joint learning across heterogeneous tasks is often treated as task coupling through feature sharing, distillation, or auxiliary supervision. However, in cross-task learning, mismatched representational and supervisory granularities make such coupling prone to interference, teacher bias, or unidirectional collapse. We argue that cross-granularity learning is fundamentally a problem of hierarchical interaction regulation rather than simple task coupling. This issue is particularly evident in UAV perception, where visual shifts and detection–segmentation objectives naturally form coarse- and fine-grained knowledge sources. To systematically study this problem, we introduce CrossUAV, a UAV benchmark for joint object detection and instance segmentation that provides a unified evaluation platform for cross-granularity task collaboration. To address these challenges, we propose Cross-Granularity Socialized Collaboration (CGSC), a progressive and adaptive framework that regulates when, where, and how tasks exchange information across network hierarchies. CGSC progressively activates cross-task interactions and adaptively adjusts the strength according to task contribution, suppressing harmful interference while exploiting complementary coarse- and fine-grained structures. Extensive experiments demonstrate consistent improvements on both tasks, validating hierarchical dynamic interaction as an effective mechanism for cross-granularity collaboration.
\end{abstract}

\section{Introduction}
\label{sec:introduction}

\begin{figure}[t]
    \centering
    \includegraphics[width=\linewidth]{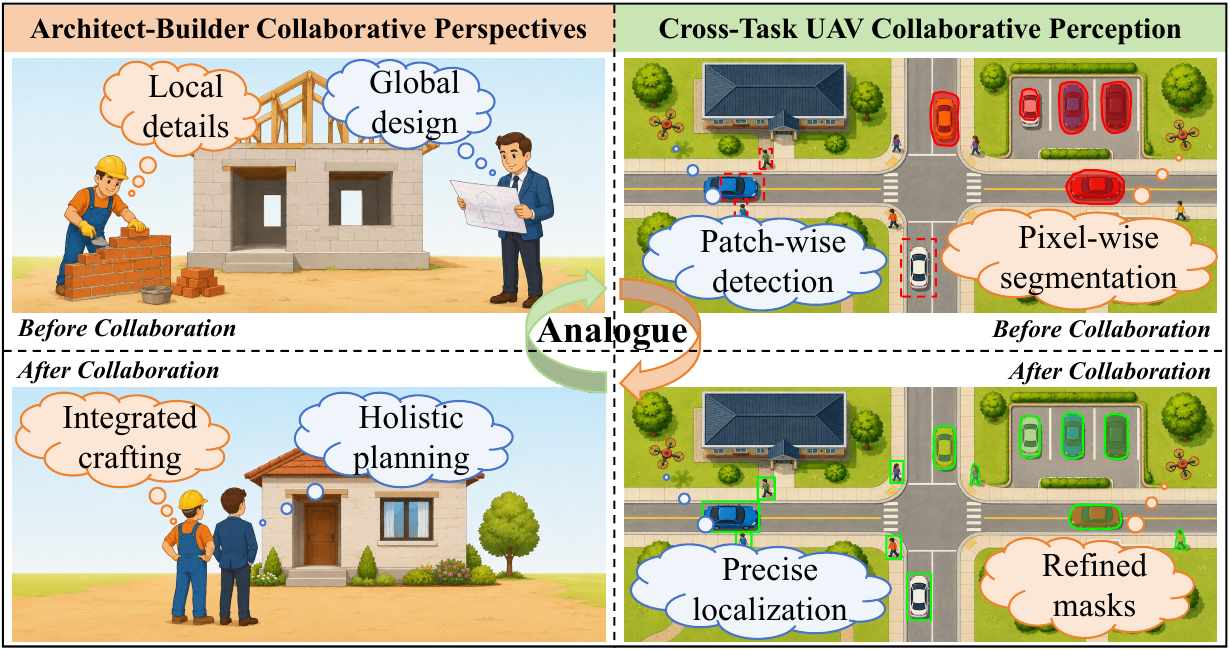}
    \caption{Inspired by human collaborative intelligence, coarse- and fine-grained tasks can mutually benefit from complementary knowledge.}
    \vskip -0.1in
    \label{fig:motivation}
\end{figure}

Inspired by human societies, intelligent agents such as UAV swarms can improve learning efficiency through collaborative knowledge exchange rather than isolated individual learning, as shown in Figure~\ref{fig:motivation}. In UAV perception, dense small objects and large viewpoint variations amplify both the structural limitations of bounding-box detection and the instability of pixel-level segmentation. Meanwhile, real-world UAV applications require not only reliable target discovery but also precise structural understanding, making detection and instance segmentation naturally complementary at different granularities. This principle underlies transfer learning and analogical learning, and has recently been formalized as Socialized Learning (SL)~\cite{yao2024socialized,li2025graphs}, which explicitly models interactions among learners. However, existing methods often lack mechanisms to account for granularity differences induced by tasks, motivating structured collaboration beyond heuristic knowledge sharing.

In machine learning, cross-task learning (CTL) relates closely to SL by explicitly modeling interactions among multiple tasks. Existing CTL methods~\cite{li2025occscene,yang2025cross} exploit task relatedness across heterogeneous objectives, outputs, or semantic granularities, but typically rely on predefined and static interaction structures, which may cause interference under granularity mismatch. A related paradigm, auxiliary learning (AL)~\cite{shin2024learning, gao2024exploiting}, introduces auxiliary tasks to improve a primary task, but its unidirectional scheme limits reciprocal knowledge exchange. These limitations suggest that cross-granularity learning requires more than task coupling.


These observations expose a key challenge in multi-model and multi-task optimization. The core issue is not merely how to connect multiple tasks or learning objectives, but how to selectively and dynamically exploit heterogeneous knowledge with different semantic granularity to improve task-specific performance without inducing negative transfer. This challenge naturally raises two questions:

\begin{enumerate}
  \item[1.] \emph{How to organize cross-granularity interaction?}
  \item[2.] \emph{How to ensure beneficial knowledge exchange?}
\end{enumerate}

Inspired by findings in cognitive science and human learning~\cite{kendal2018social}, knowledge acquisition rarely emerges from isolated practice alone. Instead, individuals improve performance through interaction with peers, selectively integrating useful information rather than indiscriminate imitation~\cite{smith2009peer,koenig2013selective}. Such interaction-driven learning enables capability improvement while preserving individual roles, allowing all participants to benefit from collaboration. Motivated by this perspective, we seek to develop a learning paradigm in which heterogeneous tasks can dynamically exchange cross-granularity knowledge, leading to mutually beneficial improvement.

To operationalize this goal, we propose a socialized cross-task learning framework that enables effective collaboration among heterogeneous tasks along three dimensions. The key issue is not merely whether tasks should interact, but where and when such interaction should be injected across the representation hierarchy.
\textbf{Who (Collaborative Participants):} collaboration is modeled among task-specialized models, each acting as a knowledgeable peer that preserves its own objective while selectively learning from complementary expertise.
\textbf{What (Cross-Granularity Knowledge):} the framework supports reciprocal exchange of task-relevant knowledge across semantic granularity, where detection and segmentation provide coarse- and fine-grained representations, respectively.
\textbf{How (Hierarchical Interaction Regulation):} a hierarchical interaction mechanism dynamically schedules collaboration and regulates information exchange across representational levels, enabling positive transfer while suppressing negative interference. We evaluate the framework on object detection and instance segmentation, demonstrating consistent and mutually beneficial performance gains. Our main contributions are summarized as follows:


\begin{itemize}
    \item We introduce CrossUAV, a UAV benchmark that reveals intrinsic granularity differences between object detection and instance segmentation.
    \item We theoretically characterize cross-granularity knowledge complementarity in socialized learning and identify conditions for positive task interaction.
    \item We propose a hierarchical interaction control framework that regulates cross-granularity interaction across both temporal dynamics and representation hierarchies.
\end{itemize}

\section{Related work}

\subsection{UAV Perception Benchmarks}
Low-altitude UAV perception is characterized by rapid viewpoint changes, dense small objects, and complex imaging conditions caused by platform motion and environmental variability. Accordingly, a number of representative benchmarks have been proposed for low-altitude object detection and segmentation.
For \textbf{detection tasks}, UAVDT~\cite{UAVDT} focuses on dense small objects and camera motion in traffic scenes, while AU-AIR~\cite{AU-AIR} and VisDrone~\cite{Visdrone} provide large-scale annotations under diverse flight conditions with severe scale variation. DroneVehicle~\cite{DroneVehicle} targets vehicle detection in low-light and nighttime scenarios, HazyDet~\cite{HazyDet} evaluates robustness under degraded visibility, and CODrone~\cite{CODrone} further extends low-altitude detection to oriented object settings.
For \textbf{segmentation tasks}, AeroScapes~\cite{AeroScapes} and SkyScapes~\cite{Skyscapes} provide fine-grained semantic annotations for urban aerial scenes, while UAVid~\cite{UAVid} and VDD~\cite{VDD} emphasize complex layouts and diverse environments. RIS-LAD~\cite{RIS-LAD} further introduces referring-expression-based drone image segmentation to enable fine-grained region understanding.


\subsection{Cross-Task Learning}
\textbf{Cross-Task Learning (CTL)} studies how signals from different tasks can facilitate each other through the interaction of tasks, rather than independent optimization. The core objective of CTL is to enable tasks to guide representation learning across heterogeneous objectives, outputs, or semantic granularity. Existing methods can be broadly categorized into two categories: \textbf{(1) Explicit knowledge transfer}~\cite{ye2020distilling,sun2021learning,xu2023multi,yao2025socialized} focuses on  sharing task outputs or intermediate features as supervisory signals. \textbf{(2) Implicit knowledge injection}~\cite{borse2023dejavu,maharana2024exposing,cao2024geometric,nishi2024joint} aims to couple tasks via joint loss or regularization without explicit knowledge transfer. For instance, X3KD~\cite{klingner2023x3kd} distills information from instance segmentation and depth estimation into multi-camera 3D object detection; BCKD~\cite{yang2023bridging} improves cross-task distillation by resolving protocol inconsistencies between classification and detection.

Existing CTL methods often use predefined, static interactions, limiting adaptation to training dynamics and causing interference across heterogeneous tasks.

\subsection{Auxiliary Learning}
\textbf{Auxiliary Learning} (AL) aims to enhance the generalization ability of a primary task by introducing auxiliary tasks during training to provide supplementary supervision beyond the target objective. Existing AL approaches can be broadly categorized according to their research focus: \textbf{(1) Signal-centric methods}~\cite{zamir2018taskonomy,chen2023joint,shin2024learning} investigate how to select, generate, or construct auxiliary tasks or supervision signals that provide informative guidance for the primary task. \textbf{(2) Optimization-centric methods}~\cite{chen2022auxiliary,chen2022module,shamsian2023auxiliary} emphasize regulating the contribution of auxiliary objectives through joint loss design, training scheduling strategies, or conflict mitigation mechanisms, to alleviate negative transfer. For example, Aux-NAS~\cite{gao2024exploiting} improves the primary task by exploiting auxiliary labels via neural architecture search, whereas ForkMerge~\cite{jiang2023forkmerge} alleviates negative transfer through adaptive optimization.

Existing AL methods are typically unidirectional, using auxiliary tasks to improve a primary task while overlooking reciprocal knowledge exchange that could benefit all.


\begin{figure*}[t]
    \centering
    \includegraphics[width=\textwidth]{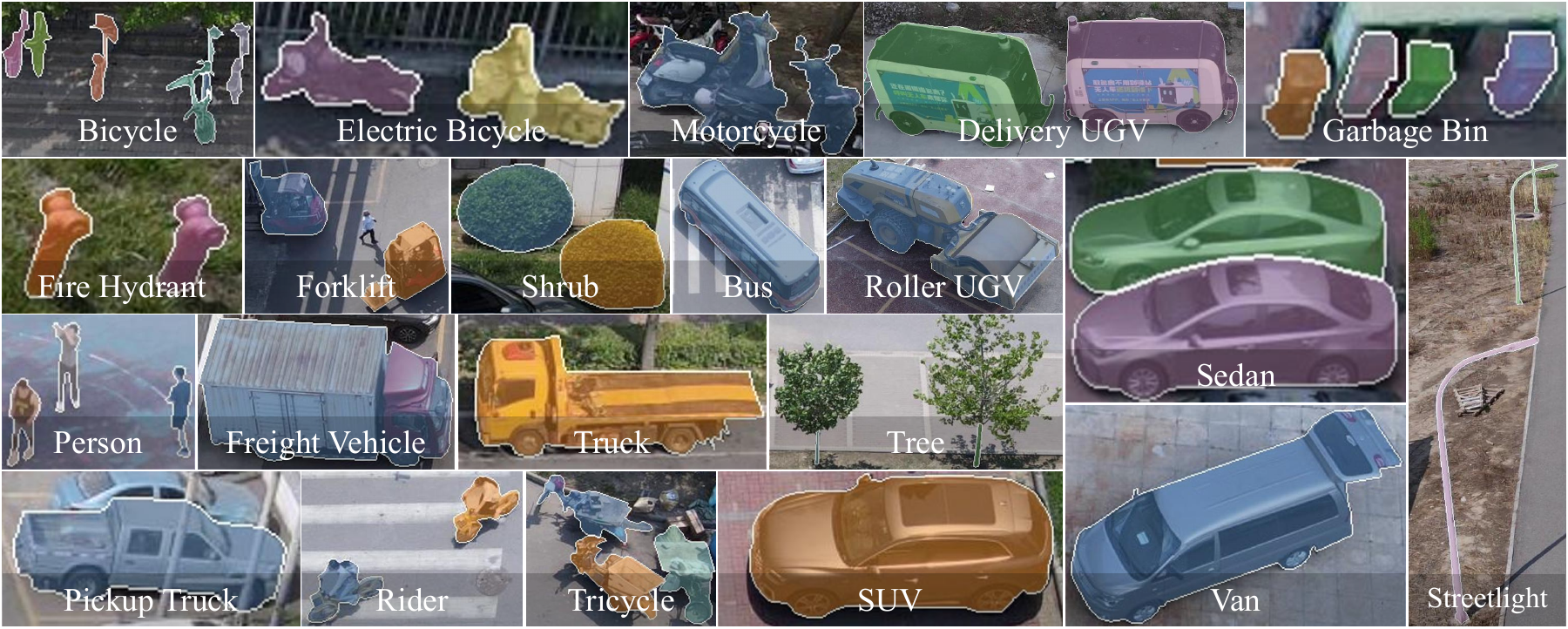}
    \caption{Representative categories and annotation examples from CrossUAV.}
    \label{fig:anno_data_examples}
    \vskip -0.1in
\end{figure*}

\begin{figure*}[t]
    \centering
    \subfloat[]{%
        \includegraphics[width=0.35\textwidth]{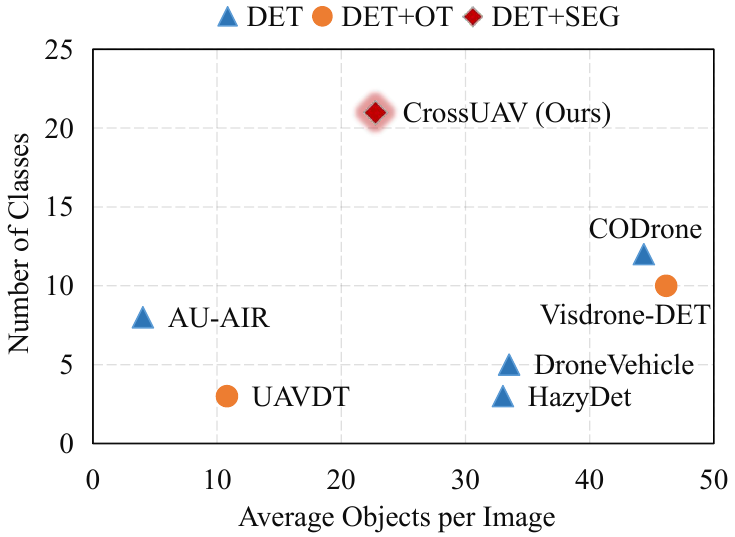}
    }
    \hfill
    \subfloat[]{%
        \includegraphics[width=0.62\textwidth]{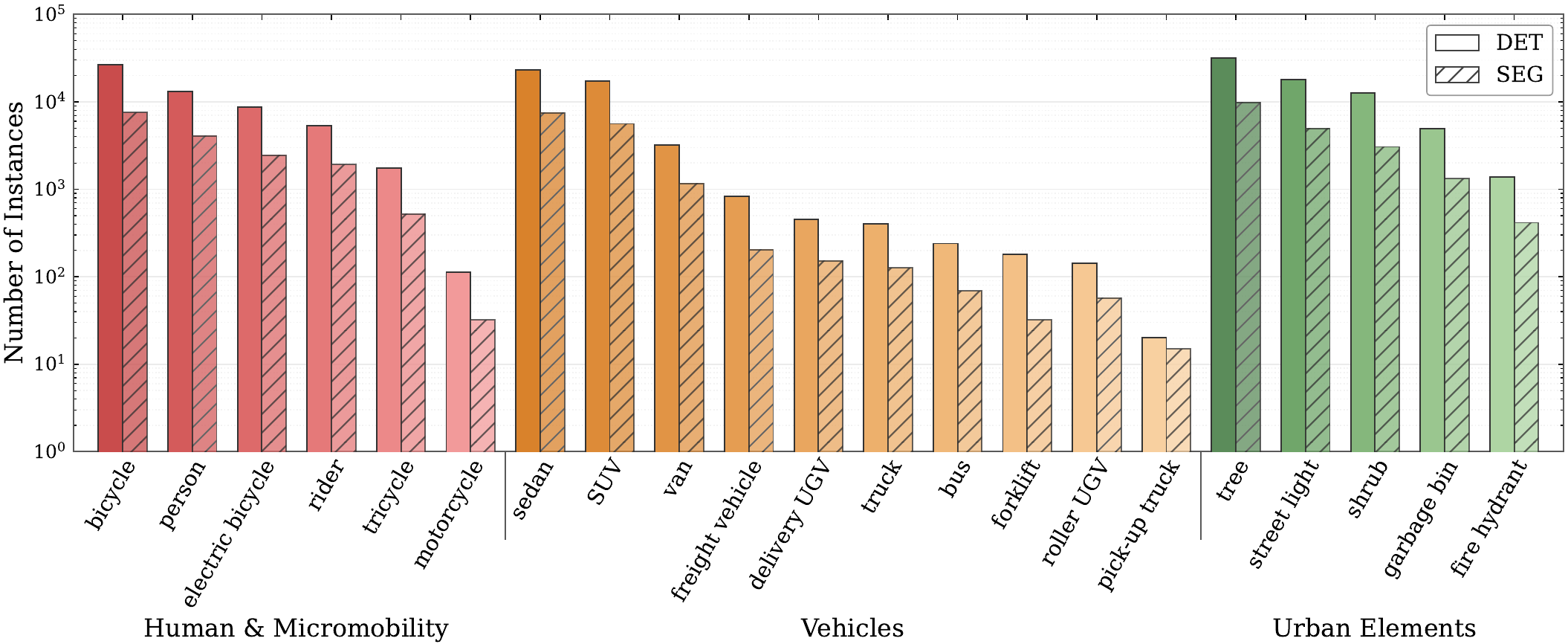}
    }

    \caption{Overview of the proposed CrossUAV dataset. (a) Comparison of CrossUAV with representative UAV benchmarks. (b) Category-wise instance statistics of CrossUAV.}
    \label{fig:overview_data}
\end{figure*}

\section{CrossUAV Benchmark}
\subsection{Benchmark Overview and Statistics}

\textbf{Dataset scale and altitude coverage.}
To facilitate cross-task interaction between object detection and instance segmentation under varying granularity, we collect aerial data using DJI Mavic Air 2 at three representative altitudes. In total, 85 video sequences comprising 143,556 RGB frames are recorded, from which 7,478 high-resolution images (1920 $\times$ 1080) are curated.

\textbf{Scene diversity and object categories.} The dataset covers diverse real-world UAV scenarios, with 21 annotated foreground object categories spanning vehicles, humans, urban facilities, and vegetation. Data are collected across multiple representative scenes, including campuses, parking lots, roads, residential areas, industrial estates, and courts. To enhance environmental diversity, data acquisition spans varying conditions, including sunny (66 sequences) and overcast (19 sequences) weather, as well as daytime (52 sequences) and dusk (28 sequences) illumination.

\subsection{Annotation Scheme}

We provide expert annotations to support both object detection and instance segmentation tasks. Specifically, we annotate object-level bounding boxes for all images and provide pixel-level instance masks for a subset of the data. Example annotations are illustrated in Figure~\ref{fig:anno_data_examples}.

\textbf{Bounding box annotation:} Object bounding boxes are annotated for all foreground categories following a consistent annotation guideline. Each object instance is labeled with a tight bounding box that closely encloses the visible region of the target. In total, the dataset contains 7,478 images with 169,958 high-quality bounding box annotations.

\textbf{Instance mask annotation:} 
Instance masks are annotated for all foreground object categories on a subset of images following a consistent annotation guideline. Each instance is labeled with a precise segmentation mask that accurately delineates the visible region of the target. In total, the dataset contains 2,247 images with 50,996 high-quality instance masks.


\subsection{Splits and Evaluation Protocol}
\textbf{Dataset splits.} CrossUAV is split into trainval and test sets at an 8:2 ratio, comprising 5,982/1,496 images for detection and 1,797/450 images for instance segmentation, respectively.

\textbf{Evaluation metrics.} Following MS COCO~\cite{coco}, we report $AP$, $AP_{50}$, and $AP_{75}$. Here, $AP$ denotes the mean precision over IoU thresholds from 0.50 to 0.95 with a step of 0.05. Detection metrics are computed using bbox IoU, while segmentation metrics use mask IoU.

\section{Cross-Granularity Collaboration in SL}
\label{sec:Cross-Granularity Collaboration in SL}

In this section, we formalize cross-granularity collaboration and analyze when knowledge exchange between coarse- and fine-grained tasks can be beneficial. Full proofs are provided in the supplementary materials.

\textbf{Problem setup.}
Let $\mathcal{M}=\{M_1,\dots,M_N\}$ denote heterogeneous models specialized for tasks $\mathcal{T}=\{T_1,\dots,T_N\}$, which share the input space $\mathcal{X}$ but differ in task-specific label spaces and information granularity. We focus on a representative pair: a coarse-grained task $T_{\mathrm{coarse}}$ such as object detection and a fine-grained task $T_{\mathrm{fine}}$ such as instance segmentation. The goal is to enable the group $\mathcal{G}=\{M_{\mathrm{coarse}},M_{\mathrm{fine}}\}$ to improve both tasks through reciprocal knowledge exchange, i.e.,
$
\mathcal{G}^{T_{\mathrm{coarse}}}>M_{\mathrm{coarse}}^{T_{\mathrm{coarse}}}
$
and
$
\mathcal{G}^{T_{\mathrm{fine}}}>M_{\mathrm{fine}}^{T_{\mathrm{fine}}}.
$

\begin{definition}[Stable Coarse-Graining]
\label{def:Tasks and Label Granularity}
Let $Y^{\mathrm{fine}}\in\mathcal{Y}_{\mathrm{fine}}$ and $Y^{\mathrm{coarse}}\in\mathcal{Y}_{\mathrm{coarse}}$ denote the supervision labels of $T_{\mathrm{fine}}$ and $T_{\mathrm{coarse}}$, respectively. We say that $T_{\mathrm{coarse}}$ is a coarse-grained task relative to $T_{\mathrm{fine}}$ if there exists a deterministic and non-invertible mapping $g:\mathcal{Y}_{\mathrm{fine}}\rightarrow\mathcal{Y}_{\mathrm{coarse}}$ such that $Y^{\mathrm{coarse}}=g(Y^{\mathrm{fine}})$. The mapping is stable if there exists a constant $L>0$ such that, for any $y_1,y_2\in\mathcal{Y}_{\mathrm{fine}}$ and non-negative losses $\ell_{\mathrm{fine}}$ and $\ell_{\mathrm{coarse}}$, it holds that

\begin{equation}
\label{eq:lipschitz_g_def}
\ell_{\mathrm{coarse}}\!\big(g(y_1), g(y_2)\big)
\le L\, \ell_{\mathrm{fine}}(y_1, y_2).
\end{equation}

Here, a smaller $L$ indicates that coarse-grained supervision better preserves the geometric structure of fine-grained labels.
\end{definition}

\begin{theorem}[Fine-to-Coarse Risk Transfer]
\label{thm:fine_to_coarse_risk}
Under the stable coarse-graining relationship in Definition~\ref{def:Tasks and Label Granularity}, let $\hat Y^{\mathrm{fine}}=h_{\mathrm{fine}}(X)$ and $\hat Y^{\mathrm{coarse}}=g(\hat Y^{\mathrm{fine}})$. Then
\begin{equation} 
\label{eq:risk_transfer} 
\resizebox{0.9\linewidth}{!}{$ 
\mathbb{E}\!\left[ \ell_{\mathrm{coarse}}\!\big(\hat Y^{\mathrm{coarse}}, Y^{\mathrm{coarse}}\big) \right] \le L\,\mathbb{E}\!\left[ \ell_{\mathrm{fine}}(\hat Y^{\mathrm{fine}}, Y^{\mathrm{fine}}) \right]. 
$} 
\end{equation}
\end{theorem}

Theorem~\ref{thm:fine_to_coarse_risk} shows that optimizing a fine-grained task can induce a bounded coarse-grained risk. This explains why fine-grained segmentation can provide structural constraints for coarse-grained detection when their labels are connected through a stable abstraction.

\begin{theorem}[Coarse-to-Fine Reliability Regularization]
\label{theorem:Generalization Bound}
Let $\mathcal{D}={(x_i,y_i^{\mathrm{fine}},y_i^{\mathrm{coarse}})}_{i=1}^n$ be i.i.d. samples, and let $\mathcal{H}_{\mathrm{fine}}$ be the hypothesis class for $T_{\mathrm{fine}}$. Suppose coarse-grained supervision induces a reliability weight
\begin{equation}
w(x)
=
\psi\!\left(
\ell_{\mathrm{coarse}}\!\left(h_{\phi}(x),y^{\mathrm{coarse}}\right)
\right),
\quad
\psi'(\cdot)\le 0,
\label{eq:weight_def}
\end{equation}
The weighted empirical risk of the fine-grained task is
\begin{equation}
\hat L_{\mathrm{fine}}^{\mathrm{joint}}(h_\theta)
\;=\;
\frac{1}{n}\sum_{i=1}^n 
w(x_i)\,
\ell_{\mathrm{fine}}\!\left(h_\theta(x_i),y_i^{\mathrm{fine}}\right).
\label{eq:joint_emp_fine}
\end{equation}
Then, for any $\delta\in(0,1)$, with probability at least $1-\delta$,
\begin{equation}
\label{eq:gen_bound_cov}
\resizebox{0.9\linewidth}{!}{$ 
\begin{aligned}
&L_{\mathrm{fine}}^{\mathrm{joint}}(h_\theta)
\;\le\;
\underbrace{\hat L_{\mathrm{fine}}^{\mathrm{joint}}(h_\theta)}_{\mathrm{Term\text{-}L(Empirical\;loss)}}
\;+\;
\underbrace{
\frac{\lVert \boldsymbol w\rVert_2}{n}\,\mathfrak R_n(\mathcal H_\mathrm{fine})
}_{\mathrm{Term\text{-}C(Complexity)}} \\
\;&+\;
\underbrace{\mathrm{Cov}\!\Big(
w(X),\,
\ell_{\mathrm{fine}}\!\left(h_\theta(X),Y^{\mathrm{fine}}\right)\Big)}_{\mathrm{Term\text{-}Cov(Covariance)}}
\;+\;
\sqrt{\frac{\ln(1/\delta)}{2n}} ,
\end{aligned}
$}
\end{equation}
where $\mathfrak{R}_n(\mathcal{H}_{\mathrm{fine}})$ denotes the Rademacher complexity of the fine-task loss class. If coarse- and fine-grained tasks share similar difficulty patterns, i.e., 
$ 
\mathrm{Cov}(\ell_{\mathrm{coarse}},\ell_{\mathrm{fine}})\ge 0, 
$ 
the monotonic weighting yields
$
\mathrm{Cov}(w(X),\ell_{\mathrm{fine}})\le0,
$
tightening the bound.
\end{theorem}

\begin{figure}[t]
    \centering
    \includegraphics[width=0.48\linewidth]{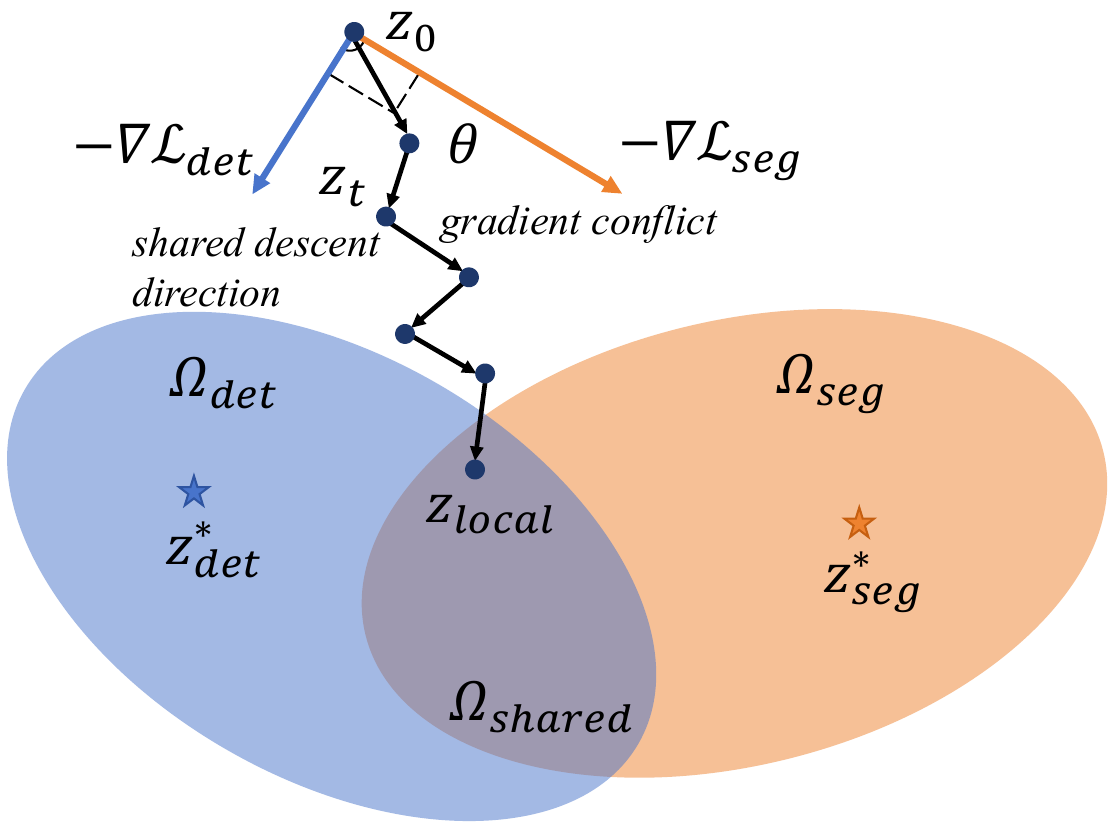}
    \includegraphics[width=0.48\linewidth]{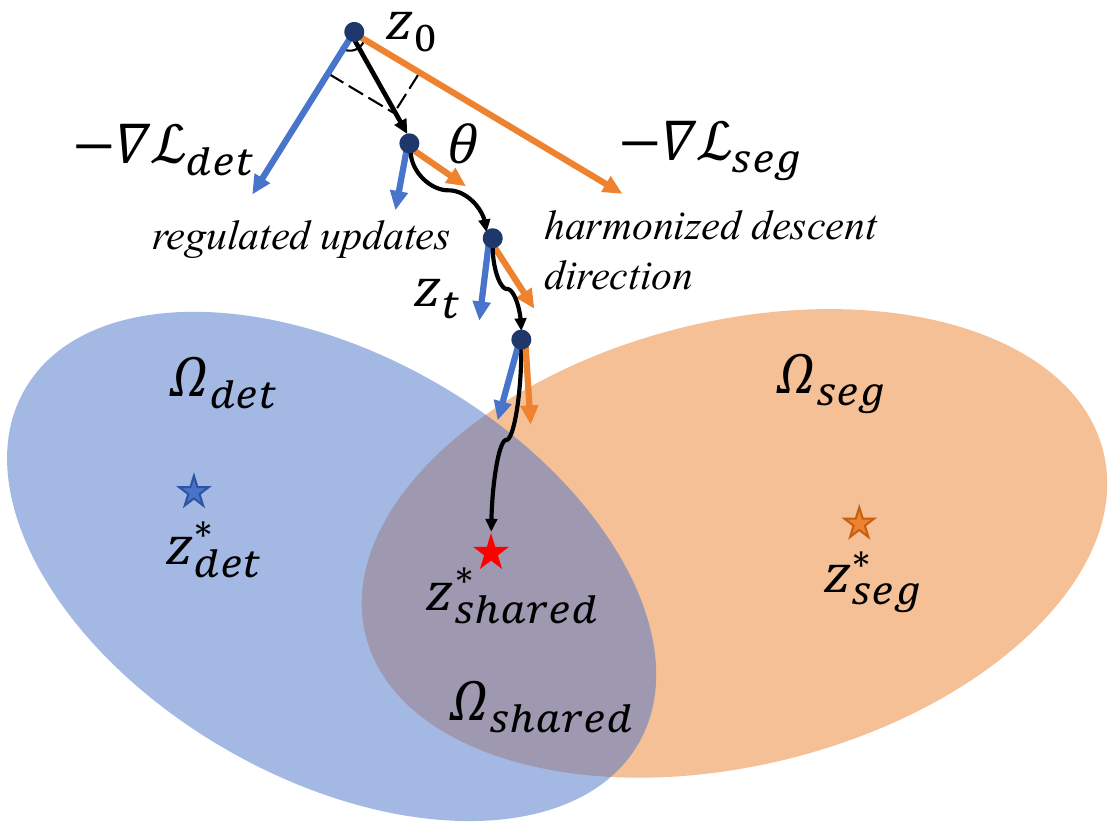}
    \caption{Illustration of cross-granularity gradient conflict and progressive harmonization. Direct joint optimization may cause conflicting task gradients, whereas regulated interaction progressively stabilizes cross-task optimization.}
    \label{fig:cross_granularity_conflict}
    \vskip -0.1in
\end{figure}

Theorem~\ref{theorem:Generalization Bound} indicates that coarse-grained supervision can act as a stable reliability signal for fine-grained learning. Together with Theorem~\ref{thm:fine_to_coarse_risk}, this analysis suggests that coarse- and fine-grained tasks can provide complementary benefits in opposite directions. However, such benefits require regulated interaction rather than indiscriminate sharing, motivating the controlled collaboration in CGSC, as shown in Fig.~\ref{fig:cross_granularity_conflict}.

\section{Methodology}
\begin{figure*}[t]
    \centering
    \includegraphics[width=\textwidth]{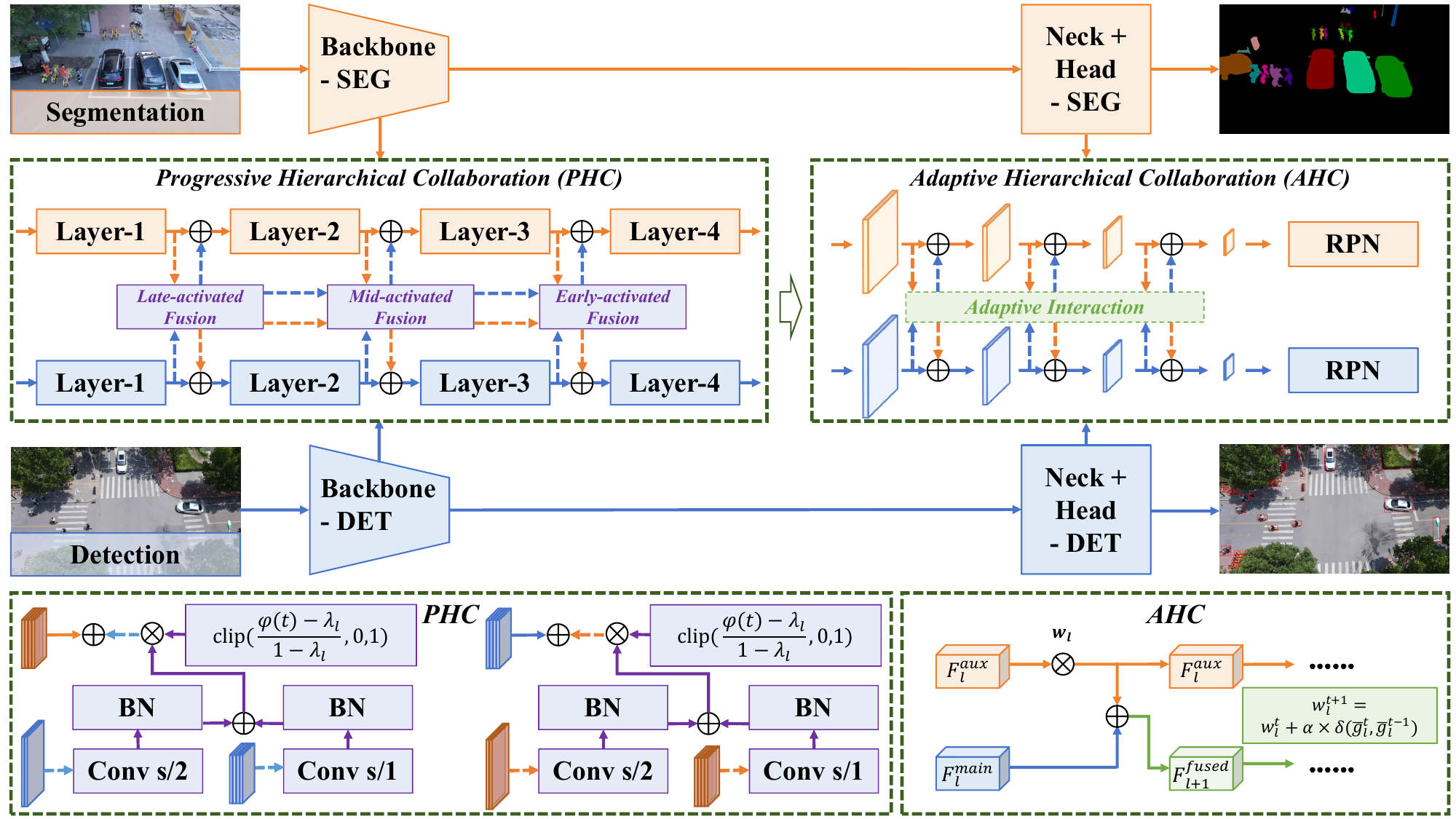}
    \caption{Overview of the proposed CGSC framework. Progressive Hierarchical Collaboration (PHC) enables progressive stage-wise interaction across backbone layers, while Adaptive Hierarchical Collaboration (AHC) dynamically regulates collaboration strength based on task contribution gains. Both enables mutual assistance between coarse- and fine-grained knowledge.}
    \label{fig:CGSC}
\end{figure*}

In this section, we introduce Cross-Granularity Socialized Collaboration (CGSC), a practical baseline for cross-task learning between object detection and instance segmentation, enabling mutual exchange between coarse- and fine-grained knowledge, as shown in Fig.~\ref{fig:CGSC}. PHC regulates the activation of cross-task interactions to stabilize representation learning, while AHC adaptively adjusts interaction strength based on task contribution gains, reinforcing beneficial interactions and suppressing negative transfer. Together, they enable stable and effective cross-task collaboration.

\subsection{The Overall Framework}
As a structured interaction pattern, CGSC governs how heterogeneous tasks influence each other during training. 
Rather than treating cross-task collaboration as a flat or static process, CGSC is grounded in the hierarchical organization of deep neural representations~\cite{chen2020towards,allen2023backward}. 
This structural property implies that cross-task interaction should be constrained by where representations interact within the network hierarchy and when such interactions are activated. 
Accordingly, we adopt hierarchical collaboration as the fundamental principle underlying CGSC. 
The interaction between tasks can be denoted as:
\begin{equation}
\label{eq:hierarchical collaboration}
    x_l^{main} = y_{l-1}^{main} + w_{l}\;y_{l-1}^{aux},\quad l \in \mathcal H,
\end{equation}
where $x_l^{main}$ and $y_l^{main}$ denote the input and output of the main branch at layer $l$, $y_l^{aux}$ denotes the corresponding auxiliary-branch output, $\mathcal H$ denotes the interaction levels, and $w_l$ controls the interaction strength.

Revisiting the theoretical analysis in Sec.~\ref{sec:Cross-Granularity Collaboration in SL}, the risk transfer bound indicates that effective optimization of the fine-grained task induces a tighter risk bound for the corresponding coarse-grained task, while the generalization bound shows that coarse-grained supervision can tighten the generalization error of the fine-grained task when properly incorporated. 
Motivated by these insights, we adopt progressive and adaptive collaboration mechanisms within the hierarchical framework to enable effective and stable cross-granularity interaction while mitigating negative transfer.

\subsection{Cross-Granularity Socialized Collaboration}
We further elaborate on CGSC, which enables effective knowledge exchange between object detection and instance segmentation through hierarchical and dynamically regulated collaboration, rather than static cross-task fusion. PHC controls the progressive evolution of cross-task interactions during representation learning, while AHC adaptively adjusts collaboration strength according to task-level contribution gains. Together, they constitute the core of the proposed socialized collaboration paradigm.

\textbf{Progressive hierarchical collaboration:} Feature representations at different hierarchical levels exhibit varying semantic stability and learning difficulty~\cite{cagnetta2024deep}. In particular, low-level features are dominated by local and task-specific patterns and are thus highly sensitive to cross-task perturbations when high-level semantics are not yet stabilized. Enforcing strong low-level interactions too early can therefore induce representation misalignment, resulting in semantic drift and unstable optimization.

To address this issue, inspired by Maslow’s Hierarchy of Needs~\cite{maslow}, PHC adopts a progressive collaboration strategy in which cross-task interactions are gradually introduced from higher to lower layers as representations become more stable. The progressive interaction strength $w_{l}$ is defined as:
\begin{equation}
\label{eq:phc_weight}
    w_l(t) = \mathrm{clip}\left(\frac{\varphi(t)-\lambda_l}{1-\lambda_l},0,1\right),
\end{equation}
where $\varphi(t)=E_n/{E_\mathrm{loops}}$ controls the overall progression of collaboration, $E_n$ denotes the current training epoch number and $E_{loop}$ is the total number of training epochs. $\lambda_l\in[0,1)$ is the interaction threshold for the $l$-th layer. Specifically, PHC constrains the influence of auxiliary-task information at each backbone layer through layer-specific thresholds, preventing auxiliary supervision from overwhelming task-specific representations while still enabling effective knowledge transfer.







\textbf{Adaptive hierarchical collaboration:} 
AHC focuses on neck-level collaboration by adaptively modulating interaction strength based on task contribution gains. Specifically, AHC introduces a feedback-driven mechanism that selectively reinforces beneficial interactions while suppressing negative transfer during optimization. Recalling the hierarchical interaction defined in Eq.~(\ref{eq:hierarchical collaboration}), AHC treats the interaction strength $w_l$ as an adaptive variable. To assess the contribution of auxiliary information at hierarchy $l$, we define a task contribution gain:
\begin{equation}
\label{eq:ahc_gain}
    \Delta_l^{t} = -\left\langle \frac{\partial \mathcal L}{\partial x_l^{\mathrm{main}}},\;y_{l-1}^{\mathrm{aux}}\right\rangle ,
\end{equation}
where $\mathcal L$ denotes the main-task objective. A positive gain indicates alignment between auxiliary features and the descent direction of the main task, whereas a negative value suggests potential interference.

Based on the temporal change of contribution gains, the interaction weight is updated as
\begin{equation}
\label{eq:ahc_update}
    w_l^{t+1} = w_l^{t}+\alpha \cdot \delta\!\left(\Delta_l^{t}, \Delta_l^{t-1}\right),
\end{equation}
where $\alpha$ is the adaptation rate and $\delta(\cdot)$ indicates the update direction. 






CGSC integrates PHC at the backbone level and AHC at the neck level, and is optimized in an end-to-end manner through a unified instance-level objective. Specifically, the overall training loss is defined as:
\begin{equation}
\label{eq:overall_loss}
\begin{aligned}
    \mathcal{L}_{\mathrm{overall}}
    &=
    \mathcal{L}_{\mathrm{cls}}
    \!\left(
    y^{\mathrm{cls}},\,
    f^{\mathrm{AHC}}\!\left(
    f^{\mathrm{PHC}}(x)
    \right)
    \right)
    \\
    &+
    \mathcal{L}_{\mathrm{bbox}}
    \!\left(
    y^{\mathrm{bbox}},\,
    f^{\mathrm{AHC}}\!\left(
    f^{\mathrm{PHC}}(x)
    \right)
    \right)
    \\
    &+
    \mathcal{L}_{\mathrm{mask}}
    \!\left(
    y^{\mathrm{mask}},\,
    f^{\mathrm{AHC}}\!\left(
    f^{\mathrm{PHC}}(x)
    \right)
    \right),
\end{aligned}
\end{equation}
where $x$ denotes the input image, and $y^{\mathrm{cls}}$, $y^{\mathrm{bbox}}$, and $y^{\mathrm{mask}}$ represent the ground-truth annotations for classification, box regression, and mask prediction, respectively. Both $y^{\mathrm{cls}}$ and $y^{\mathrm{bbox}}$ are utilized for object detection, and all losses are employed for instance segmentation.

\section{Experiments}

We conduct experiments to evaluate the proposed CGSC on the CrossUAV benchmark. We compare our approach with state-of-the-art methods. All experiments are implemented in PyTorch and conducted on 8 NVIDIA RTX 3090 GPUs.

\begin{table*}[t]
\centering
\caption{Comparison of different methods on CrossUAV under 10\% and 100\% training settings. The $1^{st}$/$2^{nd}$ best results are highlighted in \textbf{bold} or \underline{underlined}, respectively.}
\label{tab:performance comparison}
\begin{tabular}{cl l ccc ccc}
\toprule
\multirow{2}{*}{Data Setting} &
\multirow{2}{*}{Category} &
\multirow{2}{*}{Method} &
\multicolumn{3}{c}{DET (\%)} &
\multicolumn{3}{c}{SEG (\%)} \\
\cmidrule(lr){4-6} \cmidrule(lr){7-9}
& & 
& $AP$ & $AP_{50}$ & $AP_{75}$
& $AP$ & $AP_{50}$ & $AP_{75}$ \\
\midrule

\multirow{8}{*}{10\%}
& \multirow{4}{*}{Task-specific}
& RF-Next~\cite{RF-Next}
& 41.9 & 64.9 & 47.3
& 30.6 & 49.5 & 32.1 \\

& 
& RTMDet / RTMDet-Ins~\cite{RTMDet}
& 40.9 & 56.7 & 45.9
& 28.2 & 46.8 & 30.7 \\

&
& YOLO-MS~\cite{YOLO-MS}
& 46.6 & 70.9 & 50.5
& 30.1 & 52.3 & 33.8 \\

&
& RF-DETR~\cite{rfdetr}
& \underline{49.9} & \underline{71.7} & \underline{54.8}
& \textbf{36.5} & \underline{53.1} & \textbf{39.1} \\

\cmidrule(lr){2-9}

& \multirow{4}{*}{Knowledge-guided}
& INTERN~\cite{INTERN}
& 42.7 & 65.7 & 48.8
& 23.2 & 45.8 & 24.9 \\

&
& CrossKD~\cite{crosskd}
& 40.6 & 61.2 & 45.9
& 26.1 & 45.6 & 27.4 \\

&
& DISC~\cite{yao2025socialized}
& 46.5 & 67.3 & 53.6
& 29.6 & 50.3 & 32.9 \\

&
& CGSC (Ours)
& \textbf{51.2} & \textbf{73.1} & \textbf{59.1}
& \underline{35.0} & \textbf{56.4} & \underline{38.3} \\

\midrule

\multirow{8}{*}{100\%}
& \multirow{4}{*}{Task-specific}
& RF-Next~\cite{RF-Next}
& 60.2 & 85.2 & 70.0
& \underline{45.4} & \underline{70.8} & \underline{49.4} \\

&
& RTMDet / RTMDet-Ins~\cite{RTMDet}
& 64.6 & 83.9 & 72.9
& 38.3 & 61.5 & 39.8 \\

&
& YOLO-MS~\cite{YOLO-MS}
& 63.2 & 85.2 & 71.9
& 41.6 & 66.9 & 45.7 \\

&
& RF-DETR~\cite{rfdetr}
& 61.2 & \textbf{86.7} & 67.3
& 40.5 & 67.1 & 44.3 \\

\cmidrule(lr){2-9}

& \multirow{4}{*}{Knowledge-guided}
& INTERN~\cite{INTERN}
& 60.5 & 83.7 & 71.1
& 41.7 & 65.3 & 46.0 \\

&
& CrossKD~\cite{crosskd}
& \underline{63.3} & 81.0 & \underline{73.6}
& 43.4 & 67.5 & 46.7 \\

&
& DISC~\cite{yao2025socialized}
& 59.7 & 84.3 & 70.1
& 42.1 & 67.2 & 48.2 \\

&
& CGSC (Ours)
& \textbf{66.0} & \underline{85.5} & \textbf{76.5}
& \textbf{47.9} & \textbf{72.7} & \textbf{51.8} \\

\bottomrule
\end{tabular}
\end{table*}

\subsection{Experimental Setup}

\textit{Baselines:} We compare CGSC against task-specific detection and segmentation methods, as well as knowledge-guided methods based on interaction or distillation. This allows us to evaluate both its task-level effectiveness and its advantage over existing knowledge transfer paradigms.


\textit{Experimental settings:} To assess the generality of the proposed knowledge interaction mechanism, we adopt ResNet-50 as a unified backbone within a two-stage framework, which supports detection and can be readily extended to segmentation with a mask head. To evaluate performance under limited supervision, we train both tasks with 10\% and 100\% of the training set separately. All evaluations are conducted on the complete test set.

\subsection{Quantitative Results and Analysis}

We conduct comprehensive quantitative evaluations on CrossUAV, as shown in Table~\ref{tab:performance comparison}. The compared methods are grouped into task-specific methods and knowledge-guided methods, allowing us to evaluate the effectiveness of CGSC under different data scales and learning paradigms. Overall, CGSC demonstrates strong and well-balanced performance across both data settings and tasks. In the 10\% setting, CGSC achieves the best $AP_{50}$ on both tasks, while under full supervision it obtains the best AP results on both tasks, reaching 66.0 and 47.9 $AP$, respectively. This indicates that regulated cross-granularity collaboration benefits both coarse-grained localization and fine-grained mask prediction.

\textbf{Task-specific methods are limited in exploiting cross-task complementarity:}
Task-specific methods improve individual models through dedicated architectural designs, but they optimize each task independently and thus overlook the complementary relationship between coarse-grained and fine-grained supervision. In contrast, CGSC exploits cross-granularity complementarity between heterogeneous tasks: fine-grained segmentation imposes structural constraints that benefit localization, while coarse-grained detection provides a stable reliability signal to regulate mask learning.

\textbf{Knowledge transfer methods are constrained by transfer stability or teacher capacity:}
Existing knowledge-guided methods introduce additional supervision or task interaction, but their effectiveness can be limited by unstable transfer under heterogeneous task distributions. In particular, KD-based methods bound the student by the teacher's representational capacity. CGSC addresses this limitation through controllable knowledge exchange, allowing tasks to mutually refine each other without relying on a fixed supervisor.

\subsection{Ablation Study}

To further examine the effectiveness of each collaboration module in CGSC, including PHC and AHC, we perform an ablation study on CrossUAV, as reported in Table~\ref{tab:cgsc_ablation}. 

\begin{table}[htbp]
\centering
\setlength{\tabcolsep}{3pt}
\caption{Ablation study of CGSC on CrossUAV. ``\checkmark" denotes CGSC with this module.}
\label{tab:cgsc_ablation}
\begin{tabular}{ccc  ccc | ccc}
\toprule
\multicolumn{1}{c}{\multirow{2}{*}{Method}} &
\multicolumn{1}{c}{\multirow{2}{*}{PHC}} &
\multicolumn{1}{c}{\multirow{2}{*}{AHC}} &
\multicolumn{3}{c}{DET (\%)} &
\multicolumn{3}{c}{SEG (\%)} \\
\cmidrule(lr){4-6} \cmidrule(lr){7-9}
 &  &  &
\multicolumn{1}{l}{$AP$} &
\multicolumn{1}{l}{$AP_{50}$} &
\multicolumn{1}{l}{$AP_{75}$} &
\multicolumn{1}{l}{$AP$} &
\multicolumn{1}{l}{$AP_{50}$} &
\multicolumn{1}{l}{$AP_{75}$} \\
\midrule

\multirow{4}{*}{CGSC}
 &  &  &
 48.5 & 71.8 & 56.3 &
 32.1 & 54.1 & 35.1 \\

 & \checkmark &  &
 49.7 & 72.4 & 57.3 &
 33.9 & 55.6 & 36.9 \\

 &  & \checkmark &
 49.2 & 72.1 & 57.4 &
 32.8 & 54.9 & 36.1 \\

 & \checkmark & \checkmark &
 \textbf{51.2} & \textbf{73.1} & \textbf{59.1} &
 \textbf{35.0} & \textbf{56.4} & \textbf{38.3} \\

\bottomrule
\end{tabular}
\end{table}

\textbf{Without PHC and AHC:} Removing PHC and AHC degenerates CGSC to a static interaction scheme, where cross-task features are aggregated with fixed weights. This static mechanism fails to modulate cross-task interaction strength in response to task-specific granularity and scale variations, and may even introduce negative transfer between tasks.

\textbf{With PHC:} PHC adopts a stage-wise training strategy that progressively activates cross-task interactions across layers. By regulating layer-wise interaction strength via controllable thresholds, PHC regulates cross-task information flow hierarchically, facilitating effective exchange of coarse- and fine-grained representations across tasks.

\textbf{With AHC:} AHC modulates interaction strength based on task contribution gains, reinforcing beneficial interactions while suppressing interference. This adaptive regulation enables cross-task fusion to respond to task-specific contributions, leading to a more stable learning process.

\textbf{Combining PHC and AHC:} PHC and AHC jointly model progressive, layer-wise regulation at the backbone and adaptive, gain-driven interaction at the neck. PHC controls when and where cross-task interactions occur, while AHC adjusts interaction strength during optimization, enabling complementary and effective collaboration.

\begin{figure}[t]
    \centering
    \includegraphics[width=0.95\linewidth]{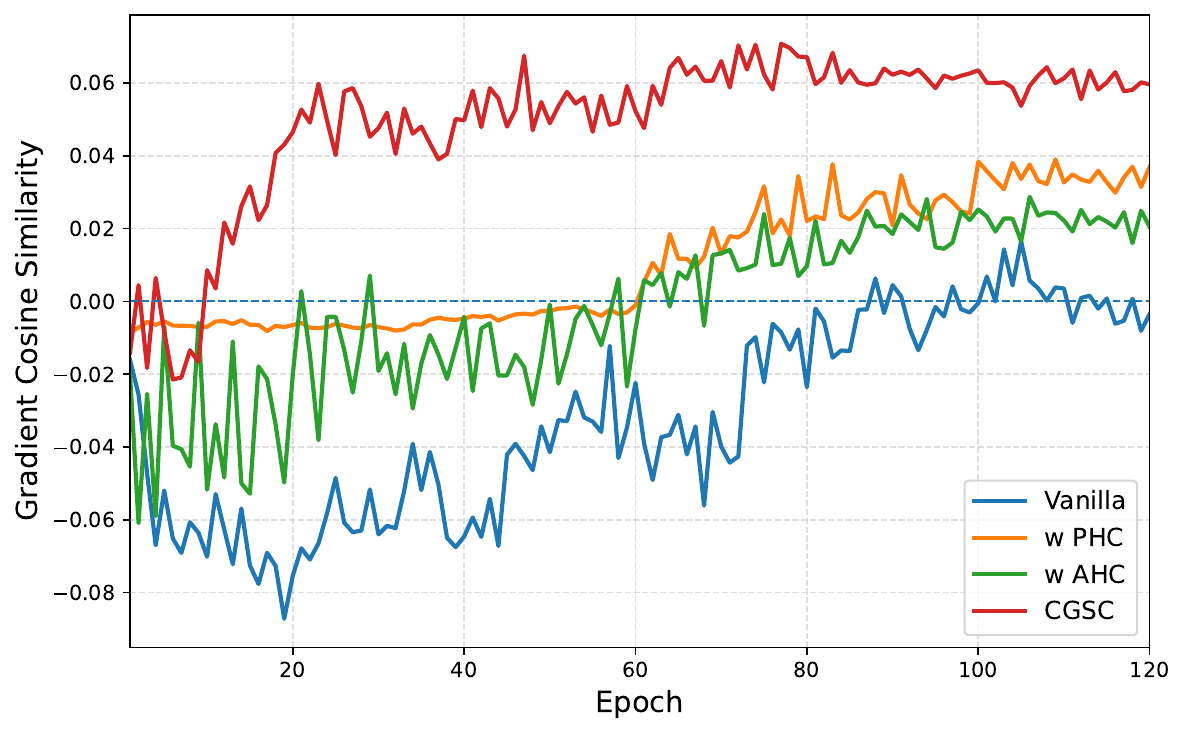}
    \caption{Gradient cosine similarity between task branches under different ablation settings.}
    \label{fig:gradient_cosine_ablation}
    \vskip -0.1in
\end{figure}


To further investigate how CGSC affects optimization, we visualize the gradient cosine similarity between the detection and segmentation branches in Fig.~\ref{fig:gradient_cosine_ablation}. Vanilla exhibits predominantly negative or weakly aligned gradients, indicating persistent competition between the two objectives. PHC mitigates this conflict by controlling when and where cross-task information is introduced, allowing task-specific representations to stabilize before broader interaction. AHC complements PHC by dynamically adjusting the optimization influence of auxiliary information, suppressing harmful updates while reinforcing beneficial ones. Their combination produces a more positive and stable alignment trend, showing that CGSC reshapes the joint optimization trajectory toward a harmonized descent process that enables reliable mutual improvement between tasks.

\section{Conclusion}

This paper studies cross-task learning under semantic granularity shifts in UAV perception and introduces CGSC, a socialized collaboration paradigm that explicitly structures interactions between heterogeneous tasks. To support systematic evaluation, we present CrossUAV, a UAV perception benchmark jointly supporting object detection and instance segmentation. Beyond methodology, we provide a theoretical analysis that characterizes when cross-granularity knowledge exchange is complementary rather than interfering, revealing conditions under which collaboration leads to positive transfer. Guided by these insights, CGSC adopts progressive and adaptive interaction to exchange coarse- and fine-grained representations, achieving consistent improvements for both tasks while suppressing negative transfer. The optimization analysis further shows that CGSC harmonizes task-branch gradients, indicating that its benefits arise from reshaping the optimization process. Overall, our results highlight principled collaboration as a key mechanism for learning under heterogeneous supervision.



\bibliographystyle{IEEEtranN}
\bibliography{references}


\end{document}


\maketitle


\section{Overview}
\label{sec:overview}

This supplementary material provides additional theoretical and
empirical support for Cross-Granularity Socialized Collaboration
(CGSC). It complements the main paper by presenting complete
theoretical derivations, detailed descriptions of the proposed
algorithms, dataset statistics and annotation examples, implementation
details, and additional experimental analyses. The remainder of this
supplementary material is organized as follows:

\begin{itemize}
    \item In Section~\ref{sec:theoretical_proofs}, we provide the
    complete theoretical analysis and proofs of the theorems introduced
    in the main paper, further clarifying the theoretical foundations of
    cross-granularity collaboration.

    \item In Section~\ref{sec:algorithms}, we present the pseudocode of
    Progressive Hierarchical Collaboration (PHC) and Adaptive
    Hierarchical Collaboration (AHC), illustrating the progressive
    feature interaction in the backbone and the adaptive feature fusion
    in the neck.

    \item In Section~\ref{sec:datasets}, we introduce the CrossUAV
    benchmark in greater detail, including its unified category system,
    dataset splits, class-wise instance statistics, and representative
    bounding-box and instance-mask annotations.

    \item In Section~\ref{sec:implementation_details}, we describe the
    model configurations, training procedures, hyperparameter settings,
    experimental environment, and the compared task-specific and
    knowledge-guided methods.

    \item In Section~\ref{sec:further_experiments}, we conduct additional
    experiments to analyze the sensitivity of the principal PHC and AHC
    hyperparameters and evaluate the cross-dataset generalization of
    CGSC on COCO.
\end{itemize}

\section{Theoretical Proofs}
\label{sec:theoretical_proofs}

In this section, we present the complete theoretical analysis and proofs of the theorems introduced in the main paper.

\subsection{Proof of Theorem 4.2}
\paragraph{Setup.}
Let $(X, Y^{\mathrm{fine}})$ be a random pair drawn from the underlying data
distribution, where $X \in \mathcal{X}$ denotes the input and
$Y^{\mathrm{fine}} \in \mathcal{Y}_{\mathrm{fine}}$ denotes the fine-grained
supervision label.
The corresponding coarse-grained label
$Y^{\mathrm{coarse}} \in \mathcal{Y}_{\mathrm{coarse}}$ is obtained through a
deterministic and non-invertible coarse-graining mapping
$g:\mathcal{Y}_{\mathrm{fine}} \rightarrow \mathcal{Y}_{\mathrm{coarse}}$ such that
\begin{equation}
Y^{\mathrm{coarse}} = g(Y^{\mathrm{fine}}).
\end{equation}
Let $h_{\mathrm{fine}}:\mathcal{X} \rightarrow \mathcal{Y}_{\mathrm{fine}}$ denote
a predictor trained for the fine-grained task, and define the induced
coarse-grained predictor as
\begin{equation}
\hat Y^{\mathrm{fine}} = h_{\mathrm{fine}}(X),\;
\hat Y^{\mathrm{coarse}} = g(\hat Y^{\mathrm{fine}}).
\end{equation}
The expected prediction risks for the fine- and coarse-grained tasks are
measured by the non-negative loss functions
$\ell_{\mathrm{fine}}$ and $\ell_{\mathrm{coarse}}$, respectively.

\paragraph{Stability of the coarse-graining mapping.}
As stated in Definition 4.1, we assume that the
coarse-graining mapping $g$ is stable with respect to the chosen loss functions,
in the sense that $g$ is $L$-Lipschitz:
\begin{equation}
\label{eq:appendix_lipschitz}
\resizebox{0.9\linewidth}{!}{$ 
\ell_{\mathrm{coarse}}\!\big(g(y_1), g(y_2)\big)
\le
L\,\ell_{\mathrm{fine}}(y_1, y_2),
\;
\forall\, y_1,y_2 \in \mathcal{Y}_{\mathrm{fine}}.
$}
\end{equation}
Such Lipschitz-type stability assumptions are standard in learning-theoretic
analyses of structured prediction and algorithmic stability
\cite{bousquet2002stability,shalev2010learnability}.

\begin{proof}
By the definition of the induced coarse-grained predictor and the supervision
hierarchy, we have
\begin{equation}
\resizebox{0.9\hsize}{!}{$
\ell_{\mathrm{coarse}}\!\big(\hat Y^{\mathrm{coarse}}, Y^{\mathrm{coarse}}\big)
=
\ell_{\mathrm{coarse}}\!\big(g(\hat Y^{\mathrm{fine}}), g(Y^{\mathrm{fine}})\big).
$}
\end{equation}
Applying the Lipschitz stability condition in
Eq.~\eqref{eq:appendix_lipschitz} yields the pointwise bound
\begin{equation}
\label{eq:appendix_pointwise}
\ell_{\mathrm{coarse}}\!\big(\hat Y^{\mathrm{coarse}}, Y^{\mathrm{coarse}}\big)
\le
L\,\ell_{\mathrm{fine}}\!\big(\hat Y^{\mathrm{fine}}, Y^{\mathrm{fine}}\big).
\end{equation}
Taking expectation over the joint distribution of $(X, Y^{\mathrm{fine}})$ on
both sides of Eq.~\eqref{eq:appendix_pointwise}, we obtain
\begin{equation}
\resizebox{0.9\hsize}{!}{$
\mathbb{E}\!\left[
\ell_{\mathrm{coarse}}\!\big(\hat Y^{\mathrm{coarse}}, Y^{\mathrm{coarse}}\big)
\right]
\le
L\,\mathbb{E}\!\left[
\ell_{\mathrm{fine}}\!\big(\hat Y^{\mathrm{fine}}, Y^{\mathrm{fine}}\big)
\right],
$}
\end{equation}
which completes the proof of Theorem 4.2.
\end{proof}

\subsection{Proof of Theorem 4.3}
\paragraph{Setup.}
Let $Z=(X,Y^{\mathrm{fine}},Y^{\mathrm{coarse}})$ be a random variable drawn from the data distribution,
and let $\mathcal D=\{Z_i\}_{i=1}^n$ be i.i.d. samples.
Define the fine-grained loss
\begin{equation}
\ell_\theta(Z)
\triangleq
\ell_{\mathrm{fine}}(h_\theta(X),Y^{\mathrm{fine}})\in[0,1],
\end{equation}
and the coarse-induced weight
\begin{equation}
\resizebox{0.95\hsize}{!}{$
    w_\theta(Z)
    \triangleq
    \psi\!\left(\ell_{\mathrm{coarse}}(h_\theta(X),Y^{\mathrm{coarse}})\right)\in[0,1],
    \quad \psi'(\cdot)\le 0.
$}
\end{equation}
The population joint fine-grained risk and its empirical counterpart are
\begin{equation}
\label{eq:joint_pop_appendix}
\begin{aligned}
L_{\mathrm{fine}}^{\mathrm{joint}}(h_\theta)
&=
\mathbb E\!\left[w_\theta(Z)\,\ell_\theta(Z)\right],\\
\hat L_{\mathrm{fine}}^{\mathrm{joint}}(h_\theta)
&=
\frac1n\sum_{i=1}^n w_\theta(Z_i)\,\ell_\theta(Z_i).
\end{aligned}
\end{equation}

We recall two standard results from statistical learning theory that will be used in the proof.

\noindent\textbf{(1) Rademacher generalization bound.}
For a class $\mathcal F$ of functions mapping $Z$ to $[0,1]$, with probability at least $1-\delta$,
\begin{equation}
\label{eq:rad_bound_appendix}
\sup_{f\in\mathcal F}\Big(\mathbb E[f(Z)]-\frac1n\sum_{i=1}^n f(Z_i)\Big)
\le
2\,\mathfrak R_n(\mathcal F)+\sqrt{\frac{\ln(1/\delta)}{2n}}.
\end{equation}
This bound is a direct consequence of the symmetrization argument and Hoeffding-type concentration
for uniformly bounded function classes, as established in the seminal work of
\citet{bartlett2002rademacher}.

\noindent\textbf{(2) Contraction principle (Lipschitz composition).}
If $\phi$ is $L$-Lipschitz and satisfies $\phi(0)=0$, then for any function class $\mathcal G$,
\begin{equation}
\label{eq:contraction_appendix}
\mathfrak R_n(\phi\circ\mathcal G)\le L\,\mathfrak R_n(\mathcal G).
\end{equation}
This inequality is a direct consequence of the contraction principle for Rademacher complexity,
which characterizes how Lipschitz transformations contract the ability of a function class to
correlate with random noise, as established in the theory of probability in Banach spaces
\citep{ledoux2013probability}.

\noindent\textbf{(3) Rademacher bound for product classes (vector contraction).}
Let $\mathcal W,\mathcal G$ be function classes mapping $Z$ to $[0,1]$.
Define the product class
$\mathcal F = \{ f(Z)=w(Z)g(Z): w\in\mathcal W, g\in\mathcal G\}$.
Then
\begin{equation}
\label{eq:rad_product}
\mathfrak R_n(\mathcal F)\le \mathfrak R_n(\mathcal W) + \mathfrak R_n(\mathcal G).
\end{equation}
\emph{Sketch.} The mapping $m(a,b)=ab$ is $1$-Lipschitz on $[0,1]^2$ w.r.t. $\ell_1$:
$|ab-a'b'|\le |a-a'|+|b-b'|$. Applying the vector contraction inequality for Rademacher complexity
to the composition $m\circ(\mathcal W,\mathcal G)$ yields~\eqref{eq:rad_product}.

\begin{proof}
Define
$f_\theta(Z)=w_\theta(Z)\ell_\theta(Z)$ and $\mathcal F=\{f_\theta:\theta\in\Theta\}$.
Since $w_\theta(Z)\in[0,1]$ and $\ell_\theta(Z)\in[0,1]$, we have $f_\theta(Z)\in[0,1]$.

By the standard Rademacher bound~\eqref{eq:rad_bound_appendix}, with probability at least $1-\delta$,
\begin{equation}
\label{eq:gen_weighted_step1}
\begin{aligned}
\sup_{\theta\in\Theta}\Bigg(
    \mathbb{E}\bigl[f_\theta(Z)\bigr]
    &- \frac{1}{n}\sum_{i=1}^{n} f_\theta(Z_i)
\Bigg) \le \\
2\,\mathfrak{R}_n(\mathcal{F}) &+ \sqrt{\frac{\ln(1/\delta)}{2n}} .
\end{aligned}
\end{equation}
Equivalently, for all $\theta\in\Theta$ simultaneously,
\begin{equation}
\label{eq:gen_weighted_step1_theta}
L_{\mathrm{fine}}^{\mathrm{joint}}(h_\theta)
\le
\hat L_{\mathrm{fine}}^{\mathrm{joint}}(h_\theta)
+
2\,\mathfrak R_n(\mathcal F)
+
\sqrt{\frac{\ln(1/\delta)}{2n}}.
\end{equation}

Introduce the two marginal classes

\begin{equation}
\begin{aligned}
    \mathcal W&=\{w_\theta(\cdot):\theta\in\Theta\}\subset[0,1], \\
    \mathcal G&=\{\ell_\theta(\cdot):\theta\in\Theta\}\subset[0,1].
\end{aligned} 
\end{equation}

Since $\mathcal F \subset \{w(\cdot)g(\cdot): w\in\mathcal W,g\in\mathcal G\}$,
by the product-class bound~\eqref{eq:rad_product},
\begin{equation}
\label{eq:radF_bound_fix}
\mathfrak R_n(\mathcal F)
\le
\mathfrak R_n(\mathcal W)+\mathfrak R_n(\mathcal G).
\end{equation}

By definition $w_\theta=\psi(\ell_{\mathrm{coarse}}(h_\theta(\cdot),y^{\mathrm{coarse}}))$.
Assume $\psi$ is $L_\psi$-Lipschitz on $[0,1]$ (bounded Lipschitz condition in the theorem),
then by the scalar contraction principle~\eqref{eq:contraction_appendix},
\begin{equation}
\label{eq:radW_fix}
\begin{aligned}
\mathfrak R_n(\mathcal W)
&\le
L_\psi \,\mathfrak R_n(\mathcal G_{\mathrm{coarse}}), \\
\mathcal G_{\mathrm{coarse}}
&=
\{\ell_{\mathrm{coarse}}(h_\theta(\cdot),y^{\mathrm{coarse}}):\theta\in\Theta\}.
\end{aligned}
\end{equation}

Similarly, if $\ell_{\mathrm{fine}}(\cdot,y^{\mathrm{fine}})$ is $L_{\mathrm{fine}}$-Lipschitz
w.r.t. the prediction (a standard assumption), then
\begin{equation}
\label{eq:radG_fix}
\begin{aligned}
    \mathfrak R_n(\mathcal G)&\le L_{\mathrm{fine}}\,\mathfrak R_n(\mathcal H), \\
    \mathfrak R_n(\mathcal G_{\mathrm{coarse}})&\le L_{\mathrm{coarse}}\,\mathfrak R_n(\mathcal H).
\end{aligned}
\end{equation}

Combining \eqref{eq:gen_weighted_step1_theta}--\eqref{eq:radG_fix} yields that,
with probability at least $1-\delta$, for all $\theta\in\Theta$,

\begin{equation}
\label{eq:gen_final_fix}
\begin{aligned}
L_{\mathrm{fine}}^{\mathrm{joint}}(h_\theta)
&\le
\hat L_{\mathrm{fine}}^{\mathrm{joint}}(h_\theta)
+
2\Bigl(
L_{\mathrm{fine}}
+
L_\psi L_{\mathrm{coarse}}
\Bigr)
\mathfrak{R}_n(\mathcal{H})
\\
&\quad
+
\sqrt{\frac{\ln(1/\delta)}{2n}} .
\end{aligned}
\end{equation}

By the covariance identity,
\begin{equation}
\label{eq:cov_identity_fix}
\begin{aligned}
\mathbb{E}\bigl[w_\theta(Z)\ell_\theta(Z)\bigr]
&=
\mathbb{E}\bigl[w_\theta(Z)\bigr]\,
\mathbb{E}\bigl[\ell_\theta(Z)\bigr]
\\
&\quad+
\operatorname{Cov}\!\bigl(
w_\theta(Z),\ell_\theta(Z)
\bigr).
\end{aligned}
\end{equation}
This decomposition is an identity and is used to interpret how the joint objective couples
the weight and the fine loss, rather than an additional inequality step.
\end{proof}

When the coarse- and fine-grained tasks share aligned semantics and exhibit similar sample-difficulty patterns, it is reasonable to assume $\mathrm{Cov}(\ell_{\mathrm{coarse}},\ell_{\mathrm{fine}})\ge 0$.
Since $\psi$ is non-increasing, the induced weight $w_\theta=\psi(\ell_{\mathrm{coarse}})$ becomes negatively associated with $\ell_{\mathrm{fine}}$, yielding $\mathrm{Cov}(w_\theta,\ell_{\mathrm{fine}})\le 0$.
This implication follows directly from classical rearrangement (Chebyshev association) inequalities for monotone transformations, as established in the theory of inequalities \citep{hardy1952inequalities}.

\section{Algorithms}
\label{sec:algorithms}
To further illustrate the implementation procedure of CGSC, we provide the pseudocode of the two core components, Progressive Hierarchical Collaboration (PHC) and Adaptive Hierarchical Collaboration (AHC). Algorithm~\ref{alg:phc} describes the progressive feature interaction  process across backbone hierarchies, while Algorithm~\ref{alg:ahc}
summarizes the adaptive feature fusion process in the neck.

\begin{algorithm}[htbp]
\caption{Progressive Hierarchical Collaboration (PHC)}
\label{alg:phc}
\begin{algorithmic}[1]
\REQUIRE Input image batch $x$
\ENSURE Enhanced hierarchical feature representations

\STATE $F_m^{0} \leftarrow x$, $F_a^{0} \leftarrow x$

\FOR{each layer $l$ in auxiliary backbone}
    \STATE $F_a^{l} \leftarrow \mathcal{B}^{(a)}_l(F_a^{l-1})$
\ENDFOR

\FOR{each layer $l$ in main backbone}
    \STATE $F_m^{l} \leftarrow \mathcal{B}^{(m)}_l(F_m^{l-1})$

    \IF{$l \in \mathcal{L}$ \textbf{and} $\omega_l > 0$}
        \STATE $\Delta_l \leftarrow \mathcal{T}_l(F_a^{l}, F_m^{l})$
        \STATE $F_m^{l} \leftarrow F_m^{l} + \omega_l \cdot \Delta_l$
    \ENDIF
\ENDFOR

\STATE \textbf{return} $\{F_m^{l}\}$
\end{algorithmic}
\end{algorithm}

\begin{algorithm}[htbp]
\caption{Adaptive Hierarchical Collaboration (AHC)}
\label{alg:ahc}
\begin{algorithmic}[1]
\REQUIRE Main backbone features $\{C_m^l\}$, auxiliary backbone features $\{C_a^l\}$ (optional)
\ENSURE Enhanced multi-level feature pyramid $\{P^l\}$

\FOR{each layer $l$}
    \STATE $L_m^l \leftarrow \text{LateralConv}_m(C_m^l)$
    \IF{auxiliary features are provided}
        \STATE $L_a^l \leftarrow \text{LateralConv}_a(C_a^l)$
    \ENDIF
\ENDFOR

\FOR{each layer $l$ from top to bottom}
    \STATE $L_m^{l-1} \leftarrow L_m^{l-1} + \text{Upsample}(L_m^{l})$
    \IF{auxiliary features are provided}
        \STATE $L_a^{l-1} \leftarrow L_a^{l-1} + \text{Upsample}(L_a^{l})$
    \ENDIF
\ENDFOR

\FOR{each layer $l$}
    \STATE $P_m^l \leftarrow \text{FPNConv}_m(L_m^l)$
    \IF{auxiliary features are provided}
        \STATE $P_a^l \leftarrow \text{FPNConv}_a(L_a^l)$
        \STATE $P^l \leftarrow P_m^l + w_l \cdot P_a^l$
    \ELSE
        \STATE $P^l \leftarrow P_m^l$
    \ENDIF
\ENDFOR

\STATE \textbf{return} $\{P^l\}$
\end{algorithmic}
\end{algorithm}

\section{Datasets}
\label{sec:datasets}
\begin{figure*}[t]
    \centering
    \includegraphics[width=\textwidth]{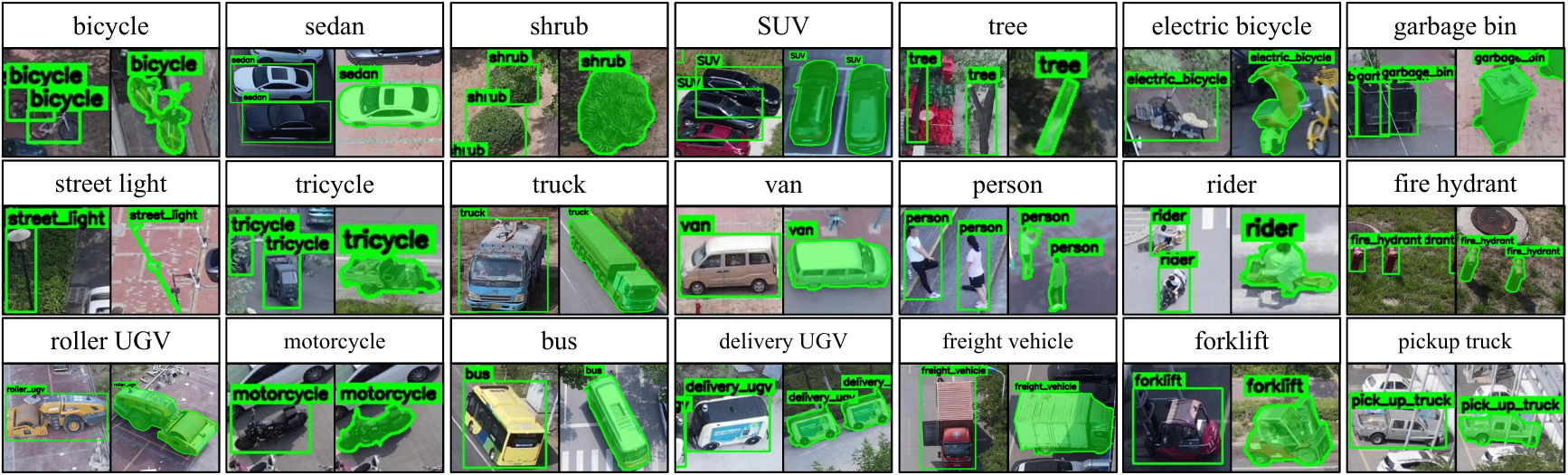}
    \caption{
    Category-wise annotation examples from CrossUAV.
    For each category, the left image presents object-level
    bounding-box annotations, while the right image shows
    pixel-level instance-mask annotations. CrossUAV provides
    a unified category space for object detection and instance
    segmentation across 21 UAV object categories.
    }
    \label{fig:crossuav_annotation_examples}
\end{figure*}

In this work, we introduce CrossUAV, a UAV perception
benchmark that jointly supports object detection and instance
segmentation under a unified category system. CrossUAV
contains 21 foreground categories covering humans,
micromobility objects, vehicles, urban facilities, and
vegetation. Representative bounding-box and instance-mask
annotations for all categories are shown in
Figure~\ref{fig:crossuav_annotation_examples}.

For object detection, the full-data setting contains 5,982
trainval images and 1,496 test images. Under the 10\%
data setting, 599 images are retained from the trainval set,
while the complete test set of 1,496 images remains unchanged.
For instance segmentation, the full-data setting contains
1,797 trainval images and 450 test images, whereas the
10\% setting contains 180 trainval images and the same
450 test images.

Detailed dataset splits and class-wise instance statistics are
reported in Table~\ref{tab:crossuav_dataset}. The instance
numbers in each row are calculated over the corresponding
trainval and test sets together. Consequently, the statistics
of the 10\% settings include instances from the reduced
trainval subset and the unchanged complete test set.

\begin{table*}[t]
\centering
\small
\setlength{\tabcolsep}{3pt}
\renewcommand{\arraystretch}{1.2}
\begin{tabularx}{\textwidth}{
    >{\raggedright\arraybackslash}p{3cm}
    >{\centering\arraybackslash}p{1.8cm}
    >{\centering\arraybackslash}p{1.8cm}
    >{\raggedright\arraybackslash}X
}
\toprule
\textbf{Dataset setting}
& \textbf{Trainval}
& \textbf{Test}
& \textbf{Class-wise instances (trainval + test)} \\
\midrule

CrossUAV-DET
& 5,982
& 1,496
&
bicycle (26,662), pick-up truck (20), sedan (23,267),
delivery UGV (451), street light (17,900), roller UGV (142),
person (13,142), electric bicycle (8,748), rider (5,332),
garbage bin (4,900), forklift (179), tricycle (1,736),
freight vehicle (826), fire hydrant (1,377), SUV (17,154),
truck (400), bus (240), shrub (12,532), motorcycle (112),
van (3,221), and tree (31,617). \\
\midrule

CrossUAV-DET (10\%)
& 599
& 1,496
&
bicycle (7,141), pick-up truck (5), sedan (6,474),
delivery UGV (116), street light (5,049), roller UGV (43),
person (3,736), electric bicycle (2,423), rider (1,478),
garbage bin (1,350), forklift (50), tricycle (488),
freight vehicle (234), fire hydrant (394), SUV (4,832),
truck (108), bus (59), shrub (3,434), motorcycle (36),
van (884), and tree (8,799). \\
\midrule

CrossUAV-SEG
& 1,797
& 450
&
bicycle (7,539), pick-up truck (15), sedan (7,454),
delivery UGV (152), street light (4,943), roller UGV (57),
person (4,078), electric bicycle (2,454), rider (1,917),
garbage bin (1,320), forklift (32), tricycle (521),
freight vehicle (203), fire hydrant (415), SUV (5,609),
truck (126), bus (69), shrub (3,056), motorcycle (32),
van (1,164), and tree (9,840). \\
\midrule

CrossUAV-SEG (10\%)
& 180
& 450
&
bicycle (2,240), pick-up truck (5), sedan (2,155),
delivery UGV (39), street light (1,333), roller UGV (16),
person (1,221), electric bicycle (704), rider (615),
garbage bin (357), forklift (21), tricycle (175),
freight vehicle (115), fire hydrant (135), SUV (1,532),
truck (28), bus (22), shrub (879), motorcycle (8),
van (371), and tree (2,704). \\
\bottomrule
\end{tabularx}
\caption{
Dataset splits and class-wise instance statistics of CrossUAV.
The 10\% setting subsamples only the trainval set, while the
complete test set is retained for evaluation. Class-wise
instance numbers are calculated over the corresponding
trainval and test sets together.
}
\label{tab:crossuav_dataset}
\end{table*}

The unified category definition ensures semantic consistency
between the detection and instance segmentation tasks.
Bounding boxes provide coarse-grained object localization,
whereas instance masks provide fine-grained structural
supervision. This annotation design enables CrossUAV to
serve as a controlled benchmark for studying reciprocal
knowledge transfer across supervision granularities.

\section{Implementation Details}
\label{sec:implementation_details}
In this section, we provide the implementation details of the models, training procedures, and datasets used in our experiments.

\subsection{Our Method}

Our proposed model is optimized using stochastic gradient descent (SGD) with a learning rate of 0.01, momentum of 0.9, weight decay of 0.0001, and Nesterov acceleration. A step-based learning rate schedule with linear warm-up for the first 500 iterations (warm-up ratio = 0.001) is adopted. Models trained on the full dataset are optimized for 24 epochs with learning rate decays at epochs 16 and 22, while models under the 10\% data setting are trained for 120 epochs with decays at epochs 80 and 110. For the Progressive Hierarchical Collaboration (PHC), we adopt a layer-wise progressive activation schedule with thresholds
$\lambda_3=0$, $\lambda_2=0.2$, and $\lambda_1=0.4$ for the hierarchical features. This setting enables higher-level features to participate in cross-task interaction earlier, while gradually introducing lower-level feature collaboration as training progresses. For the Adaptive Hierarchical Collaboration (AHC), we set the adaptation rate $\alpha=0.05$ in all experiments. The selection of the PHC and AHC hyperparameters is further discussed in the hyperparameter analysis~\ref{sec:hyperparameter analysis}.

The batch size is set to 8 for both object detection and instance segmentation tasks. Our model is implemented and deployed using Python 3.8.12 and PyTorch 1.8.0. All experiments are conducted on a machine equipped with 8 NVIDIA RTX 3090 GPUs. All reported results are obtained by averaging over three independent runs.

\subsection{Compared Methods}
In this subsection, we briefly introduce the compared methods. 
For all competing methods, we adopt their official open-source implementations with the training configurations provided by the authors. The compared methods are categorized into task-specific methods and knowledge-guided methods according to their learning paradigms.

\textbf{Task-specific methods:}
These methods optimize individual vision tasks independently without explicitly modeling cross-task interactions.

\begin{itemize}
    \item \textbf{RF-Next~\cite{RF-Next}:} 
    a CNN-based framework that revisits receptive field design for visual recognition. 
    By optimizing receptive field distribution and feature aggregation strategies, RF-Next enhances spatial context modeling and improves feature representation quality.

    \item \textbf{RTMDet~\cite{RTMDet}:} 
    a real-time object detection framework that adopts an efficient fully convolutional architecture with optimized label assignment and training strategies. 
    It achieves a favorable trade-off between detection accuracy and inference efficiency.

    \item \textbf{YOLO-MS~\cite{YOLO-MS}:} 
    a multi-scale object detection framework that focuses on improving scale-aware feature representation. 
    YOLO-MS introduces efficient multi-scale aggregation strategies to enhance object detection performance under diverse object sizes and complex visual conditions.

    \item \textbf{RF-DETR~\cite{rfdetr}:} 
    a transformer-based object detection framework that incorporates refined feature learning and attention mechanisms to improve detection accuracy. 
    RF-DETR leverages the strengths of transformer architectures for global context modeling and robust object localization.
\end{itemize}

\textbf{Knowledge-guided methods:}
These methods introduce additional knowledge transfer or interaction mechanisms to exploit relationships between different tasks or models.

\begin{itemize}
    \item \textbf{INTERN~\cite{INTERN}:} 
    a representation-centric vision learning paradigm that aims at general visual intelligence. 
    INTERN learns structured and task-agnostic representations, allowing multiple downstream tasks to be formulated as different decoding views over a shared representation space.

    \item \textbf{CrossKD~\cite{crosskd}:} 
    a cross-head knowledge distillation method for object detection.
    It passes intermediate student-head features through the teacher head and distills the resulting predictions, reducing conflicts between ground-truth supervision and teacher guidance.

    \item \textbf{DISC~\cite{yao2025socialized}:} 
    a knowledge transfer framework that enables dynamic interaction among heterogeneous models. 
    By modeling inter-model relationships during training, DISC adaptively transfers useful knowledge while reducing ineffective knowledge exchange.
\end{itemize}

\section{Further Experiments}
\label{sec:further_experiments}
This section provides additional experiments to examine the
hyperparameter sensitivity and cross-dataset generalization
of CGSC. These results complement the primary evaluations
on CrossUAV presented in the main paper.

\subsection{Hyperparameter Analysis}
\label{sec:hyperparameter analysis}

We investigate the sensitivity of CGSC to the principal
hyperparameters of Progressive Hierarchical Collaboration
(PHC) and Adaptive Hierarchical Collaboration (AHC).
Considering the computational cost of repeated training, all
hyperparameter experiments are conducted under the 10\%
training-data setting, while the complete test set is retained
for evaluation. Unless otherwise specified, all optimization
settings follow those described in the implementation details.
Each configuration is independently trained three times, and we reported the mean result.

To avoid selecting hyperparameters that favor only one task,
we use the balanced performance
\begin{equation}
\label{eq:balanced_ap}
\mathrm{AP}_{\mathrm{avg}}
=
\frac{1}{2}
\left(
\mathrm{AP}_{\mathrm{DET}}
+
\mathrm{AP}_{\mathrm{SEG}}
\right)
\end{equation}
as the primary selection criterion. Meanwhile, we require the
selected configuration to improve both detection and
segmentation, consistent with the mutually beneficial objective
of cross-granularity collaboration. When analyzing one
component, the hyperparameters of the other component are
fixed to their default values.

PHC controls when cross-task interactions are activated at
different backbone levels. Since the highest-level feature
contains relatively stable semantic information, we fix
$\lambda_3=0$ such that Layer-3 interaction starts from the
beginning of training. We then vary the activation thresholds
of the middle- and lower-level features using
\begin{equation}
\label{eq:phc_threshold_candidates}
\begin{aligned}
(\lambda_3,\lambda_2,\lambda_1)
\in \Bigl\{&
(0,0.1,0.2),\ (0,0.2,0.4),\\
&
(0,0.3,0.6),\ (0,0.4,0.8)
\Bigr\}.
\end{aligned}
\end{equation}

The adaptation rate of AHC is fixed to $\alpha=0.05$ in this
experiment.

According to Eq.~(7) in the main paper, interaction at layer
$l$ starts when the normalized training progress
$\phi(t)$ reaches $\lambda_l$. Under the 120-epoch training
schedule, the four threshold configurations therefore activate
Layers 3, 2, and 1 at epochs $(0,12,24)$, $(0,24,48)$,
$(0,36,72)$, and $(0,48,96)$, respectively. These settings
cover interaction schedules ranging from relatively aggressive
early collaboration to conservative delayed collaboration.

\begin{table}[t]
    \centering
    \small
    \setlength{\tabcolsep}{3.5pt}
    \begin{tabular}{ccccc}
        \toprule
        $(\lambda_3,\lambda_2,\lambda_1)$
        & Activation epochs
        & $AP_{\mathrm{DET}}$
        & $AP_{\mathrm{SEG}}$
        & $\mathrm{AP}_{\mathrm{avg}}$ \\
        \midrule
        $(0,0.1,0.2)$ & $(0,12,24)$
        & 50.7 & 34.3 & 42.5 \\
        $(0,0.2,0.4)$ & $(0,24,48)$
        & \textbf{51.2} & \textbf{35.0} & \textbf{43.1} \\
        $(0,0.3,0.6)$ & $(0,36,72)$
        & 50.6 & 34.6 & 42.6 \\
        $(0,0.4,0.8)$ & $(0,48,96)$
        & 50.2 & 33.4 & 41.8 \\
        \bottomrule
    \end{tabular}
    \caption{Sensitivity analysis of the PHC thresholds under the 10\% data setting. The activation epochs correspond to Layers 3, 2, and 1, respectively. Results are reported as the average over three independent runs.}
    \label{tab:phc_hyperparameter}
\end{table}

The comparison examines the trade-off between interaction
sufficiency and representation stability. Activating lower-level
interactions too early may introduce cross-task perturbations
before task-specific representations become sufficiently
stable. In contrast, excessively delayed activation leaves
limited training time for lower-level cross-task knowledge to
be effectively integrated. The final thresholds are selected
according to the balanced AP in
Table~\ref{tab:phc_hyperparameter}, while ensuring that both
tasks benefit from collaboration.

The configuration $(0,0.2,0.4)$ achieves the best balanced
performance across detection and segmentation. Compared with
the earlier schedule, it provides sufficient time for
task-specific representations to stabilize before lower-level
interaction is introduced. Meanwhile, it retains a longer
collaboration period than the more conservative schedules.

The adaptation rate $\alpha$ determines the magnitude of each
AHC interaction-weight update. A small value leads to
conservative weight adjustment and may respond slowly to
changes in auxiliary-task contribution, whereas a large value
produces more aggressive updates and may cause interaction
weights to fluctuate during optimization.

We evaluate
\begin{equation}
\alpha \in \{0.01, 0.03, 0.05, 0.10\},
\end{equation}
covering conservative, moderate, and relatively aggressive
adaptation rates. During this analysis, the PHC thresholds are
fixed to
$(\lambda_3,\lambda_2,\lambda_1)=(0,0.2,0.4)$.

\begin{table}[t]
    \centering
    \small
    \setlength{\tabcolsep}{5pt}
    \begin{tabular}{cccc}
        \toprule
        $\alpha$
        & $AP_{\mathrm{DET}}$
        & $AP_{\mathrm{SEG}}$
        & $\mathrm{AP}_{\mathrm{avg}}$ \\
        \midrule
        $0.01$ & 49.8 & 34.0 & 41.9 \\
        $0.03$ & 50.5 & 34.3 & 42.4 \\
        $0.05$ & \textbf{51.2} & \textbf{35.0} & \textbf{43.1} \\
        $0.10$ & 49.6 & 33.5 & 41.6 \\
        \bottomrule
    \end{tabular}
    \caption{Sensitivity analysis of the AHC adaptation rate under the 10\% data setting. Results are reported as the average over three independent runs.}
    \label{tab:ahc_hyperparameter}
\end{table}

As shown in Table~\ref{tab:ahc_hyperparameter}, the
adaptation rate controls the trade-off between responsiveness
and stability. Smaller values produce smoother but slower
adjustments, which may prevent AHC from responding promptly
to changes in cross-task contribution. Conversely, excessively
large values may cause abrupt variations in interaction
strength, reducing the stability of knowledge exchange. We
select the final adaptation rate according to the balanced AP
across the two tasks.

Among the evaluated values, $\alpha=0.05$ provides the best
balance between detection and segmentation. It responds
sufficiently to changes in task contribution while avoiding
the unstable interaction-weight variations observed with a
larger adaptation rate.

\subsection{Cross-Dataset Generalization}

\begin{table*}[t]
\centering
\renewcommand{\arraystretch}{1}

\begin{tabular}{lccc|ccc}
\toprule
\multirow{2}{*}{\textbf{Method}} &
\multicolumn{3}{c|}{\textbf{DET (\%)}} &
\multicolumn{3}{c}{\textbf{SEG (\%)}} \\
\cmidrule(lr){2-4}
\cmidrule(lr){5-7}
&
$AP$ & $AP_{50}$ & $AP_{75}$ &
$AP$ & $AP_{50}$ & $AP_{75}$ \\
\midrule
Faster R-CNN~\cite{faster-rcnn}
& 30.5 & 47.5 & 32.7
& 27.4 & 46.8 & 28.6 \\

INTERN~\cite{INTERN}
& 26.9 & 45.8 & 28.0
& 22.8 & 42.0 & 22.3 \\

RF-Next~\cite{RF-Next}
& 32.1 & 51.0 & 34.6
& 24.8 & 44.6 & 24.7 \\

RTMDet~\cite{RTMDet}
& 35.0 & 53.9 & 38.2
& \underline{28.4} & \underline{48.9}
& \underline{29.7} \\

CrossKD~\cite{crosskd}
& 28.1 & 47.3 & 29.6
& 23.9 & 43.4 & 23.8 \\

DISC~\cite{yao2025socialized}
& \underline{35.7} & \underline{54.5}
& \underline{39.1}
& 28.2 & \underline{48.9} & 29.4 \\
\midrule

CGSC (Ours)
& \textbf{36.6} & \textbf{55.6} & \textbf{40.2}
& \textbf{29.5} & \textbf{50.0} & \textbf{31.2} \\
\bottomrule
\end{tabular}

\caption{Cross-dataset generalization results on COCO under
the data-efficient setting. The $1^{st}$/$2^{nd}$ best results are highlighted in \textbf{bold} or \underline{underlined}, respectively.}
\label{tab:coco_comparison}
\end{table*}

Although CGSC is primarily developed for cross-granularity
collaboration in UAV perception, we further evaluate it on
COCO to examine whether the proposed interaction mechanism
generalizes beyond the CrossUAV benchmark. This experiment
serves as an additional validation of the method rather than
the primary benchmark evaluation, as the main focus of this
work is cross-task learning under low-altitude UAV imagery. 
As shown in Table~\ref{tab:coco_comparison}, CGSC achieves
the best results across all detection and instance segmentation
metrics. For object detection, CGSC obtains 36.6 AP,
outperforming the strongest competing method DISC by
0.9 AP. For instance segmentation, CGSC reaches 29.5 AP,
surpassing the best task-specific baseline by 1.1 AP.
Consistent improvements are also observed on AP$_{50}$
and AP$_{75}$ for both tasks.

These results indicate that the progressive and adaptive
interaction mechanisms of CGSC are not restricted to the
specific visual distribution of CrossUAV. Nevertheless,
CrossUAV remains the primary benchmark of this work, as it
provides a unified UAV platform for studying collaboration
between coarse-grained detection and fine-grained instance
segmentation under dense objects, viewpoint variations, and
low-altitude imaging conditions.


\bibliographystyle{IEEEtranN}
\bibliography{references}